%% file: paper.tex
\documentclass[a4paper]{article}

\usepackage{INTERSPEECH2020}
\usepackage[dvipsnames]{xcolor}
\usepackage{amsfonts}
\usepackage{tipa}
\usepackage{cite}
\usepackage{microtype}
\usepackage{enumitem}
\usepackage{siunitx}
\usepackage{footnote}
\usepackage{tablefootnote}
\usepackage{float}

\makesavenoteenv{table}
\makesavenoteenv{tabular}

\newcommand{\klcomment}[1]{\textcolor{red}{#1 (KL)}\ }
\newcommand{\sscomment}[1]{\textcolor{RedOrange}{#1 (SS)}\ }
\newcommand{\yhcomment}[1]{\textcolor{teal}{#1 (YH)}\ }

\renewcommand{\klcomment}[1]{}
\renewcommand{\sscomment}[1]{}
\renewcommand{\yhcomment}[1]{}

\newcommand{\kledit}[1]{{#1}\ }

\newcommand{\goodbyewordsim}[1]{}

\title{Multilingual Jointly Trained Acoustic and Written Word Embeddings 
}

\name{Yushi Hu$^1$, Shane Settle$^2$, Karen Livescu$^2$}
\address{$^1$University of Chicago\\$^2$TTI-Chicago}
\email{hys98@uchicago.edu, \{settle.shane,klivescu\}@ttic.edu}

\begin{document}

\maketitle

\input{abstract}

\input{intro}
\input{approach}

\input{setup}

\input{results}
\input{conclusions}
\vspace{-.1in}
\section{Acknowledgements}
\vspace{-.05in}
This research was funded by NSF award IIS-1816627, and by an AWS Machine Learning Research Award.  We thank Herman Kamper for helpful feedback.
\newpage
\bibliographystyle{IEEEtran}
\bibliography{refs}

\end{document}

%% file: abstract.tex
\begin{abstract}
Acoustic word embeddings (AWEs) are vector representations of spoken word segments.  AWEs can be learned jointly with embeddings of character sequences, to generate phonetically meaningful embeddings of written words, or acoustically grounded word embeddings (AGWEs).  Such embeddings have been used to improve speech retrieval, recognition, and spoken term discovery.  In this work, we extend this idea to multiple low-resource languages. We jointly train an AWE model and an AGWE model, using phonetically transcribed data from multiple languages. The pre-trained models can then be used for unseen zero-resource languages, or fine-tuned on data from low-resource languages.  We also investigate distinctive features, as an alternative to phone labels, to better share cross-lingual information. We test our models on word discrimination tasks for twelve languages. When trained on eleven languages and tested on the remaining unseen language, our model outperforms traditional unsupervised approaches like dynamic time warping.  After fine-tuning the pre-trained models on one hour or even ten minutes of data from a new language, performance is typically much better than training on only the target-language data.  We also find that phonetic supervision improves performance over character sequences, and that distinctive feature supervision is helpful in handling unseen phones in the target language.
\end{abstract}
\noindent\textbf{Index Terms}: acoustic word embeddings, multilingual, low-resource, zero-resource, distinctive features.

%% file: intro.tex
\section{Introduction}

Acoustic word embeddings (AWEs) are vector representations of spoken word segments of arbitrary duration~\cite{levin+etal_asru13}.  AWEs are an attractive tool in tasks involving reasoning about whole word segments, as they provide a compact representation of spoken words and can be used to efficiently measure similarity between segments.  For example, AWEs have been used to speed up and improve query-by-example  search~\cite{levin+etal_icassp15,settle2017query,yuan2018learning}, where the distance computation traditionally done with dynamic time warping or subword methods is replaced by vector distances between embeddings.  AWEs have also been used for unsupervised segmentation and spoken term discovery~\cite{kamper2016unsupervised}, where the use of embeddings can eliminate the need to explicitly model subword units.

A variety of approaches have been explored for constructing and learning AWEs, including template-based techniques~\cite{levin+etal_asru13}, discriminative neural network models~\cite{Kamper_16a,settle2016discriminative}, and unsupervised autoencoder-based models~\cite{kamper2018truly,kamper2020multilingual,audhkhasi2017end,chung2016unsupervised,Holzenberger2018}.  Despite the flexibility of fully unsupervised methods, supervised AWE training can be highly data-efficient, providing large performance improvements over unsupervised methods on downstream tasks with just 100 minutes of labelled audio~\cite{settle2017query}.

\begin{figure}[t]
  \centering
  \includegraphics[width=0.8\linewidth]{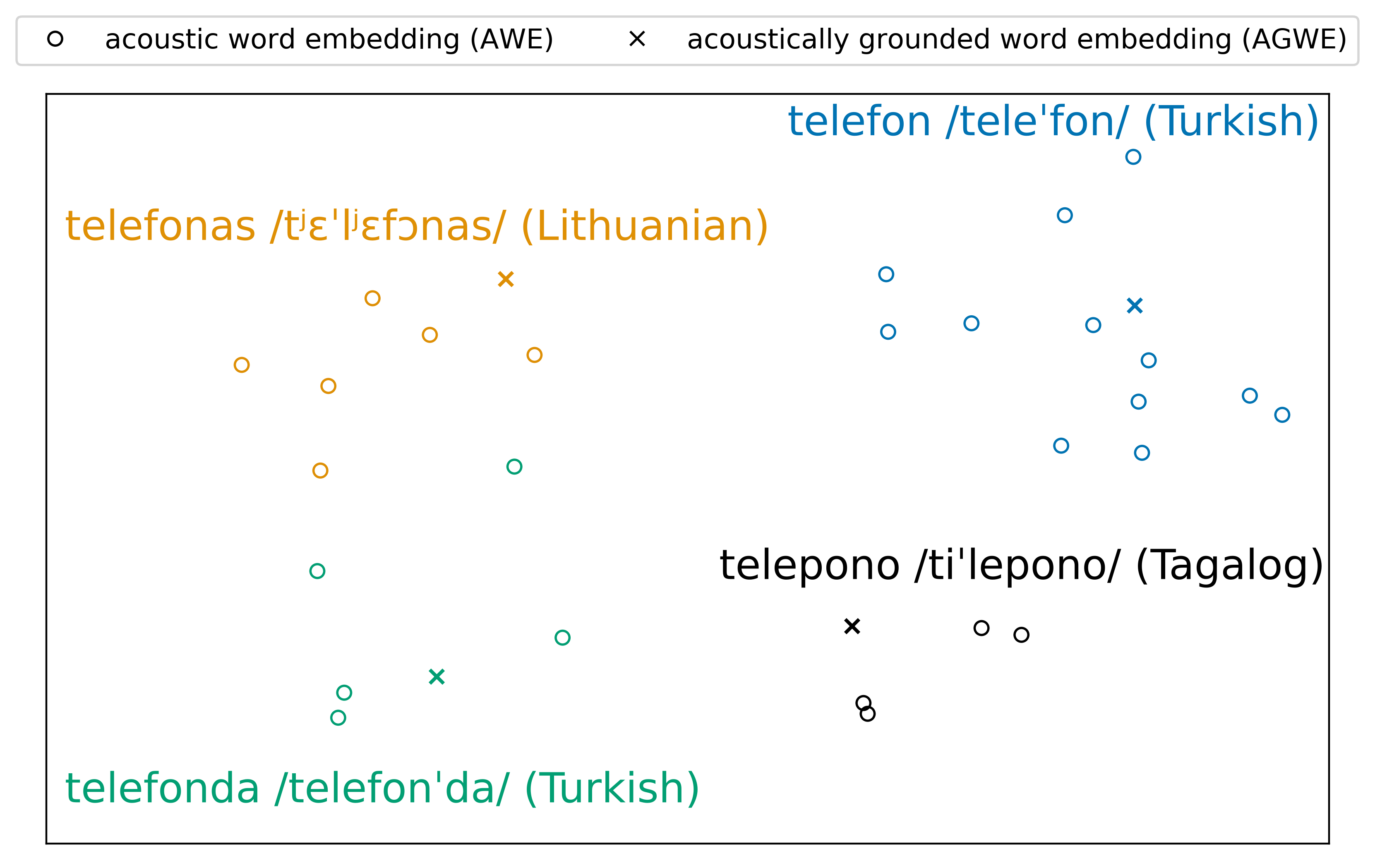}
  \vspace{-0.3cm}
  \caption{Examples of jointly trained acoustic and written word embeddings, for the Turkish word ``telefon" and several of its nearest neighbors, visualized with t-SNE~\cite{MaatenHinton08a}.
  The acoustic word embeddings (AWEs) are clustered around the corresponding acoustically grounded written word embeddings (AGWEs). 
  }
\label{fig:words-tsne}
\vspace{-0.3cm}
\end{figure}

Some tasks involve measuring similarity between spoken and written words.  For such purposes it can be useful to learn both embeddings of spoken words and embeddings of written words that represent the word's phonetic content, which we refer to as {\it acoustically grounded word embeddings} (AGWEs).
For example, AWEs and AGWEs have been used together for spoken term detection~\cite{audhkhasi2017end}, where a written query word can be compared to speech segments directly via vector similarity between their embeddings; and to whole-word speech recognition, where such embeddings have been used either for rescoring~\cite{BengioHeigol14a} or to initialize an end-to-end neural model and improve recognition of rare and out-of-vocabulary words~\cite{settle2019acoustically}.
AWEs and AGWEs can be learned jointly~\cite{BengioHeigol14a,he+etal_iclr2017} to embed spoken and written words in the same vector space (Figure~\ref{fig:words-tsne}). \klcomment{I removed a reference to palaskar -- i think it's OK to leave it out. also added references here to bengio and heigold and rephrased accordingly}

Prior work in this area has largely focused on English.  In this work, we study the learning of AWEs/AGWEs for multiple languages, in particular low-resource languages.  Recent related work~\cite{kamper2020multilingual} has begun to explore multilingual AWEs, specifically for zero-resource languages, including unsupervised approaches trained on a zero-resource language of interest and \kledit{supervised models} trained on multiple additional languages and applied to a zero-resource language.  Our work complements this prior work by exploring, in addition to the zero-resource regime, a number of low-resource settings, and the trade-off between performance and data availability.
In addition, we learn not only AWEs but also AGWEs, thus widening the range of tasks to which our models apply.  One product of our work is a set of written word embeddings for multiple languages in the same phonetically meaningful embedding space. 

Our approach extends prior work on AWE+AGWE training for English, based on joint embedding training with a multi-view contrastive loss~\cite{he+etal_iclr2017,settle2019acoustically}.  In contrast with prior work, we use phonetic pronunciations as supervision rather than characters,
to avoid the difficulties introduced by multiple
written alphabets.
In addition, to 
better model rare or unseen phones, we explore the use of distinctive features as an alternative. We evaluate our embeddings by their performance on word discrimination, a measure that is independent of any particular downstream task.

Our main finding is that, in low-resource settings, pre-training embeddings on multiple languages and fine-tuning on a small amount of target language data produces much higher-quality embeddings than training on target language data alone.  In addition, performance is improved by using a pronunciation dictionary to embed the phone sequences of written words, rather than using the character sequence as in prior work.  We also find that using distinctive features in place of phones has advantages in settings where there are unseen phones in the target language.  We provide results showing how embedding quality changes with increasing training set size, and in particular find that modest data sizes
($\sim10$ hrs)
are likely sufficient for training AWEs/AGWEs with our approach.

%% file: approach.tex
\vspace{-.05in}
\section{Embedding Models}
\vspace{-.05in}
An acoustic word embedding (AWE) model $f$ maps a variable-length spoken segment ${\bf X} \in \mathbb{R}^{T \times D}$, where $T$ is the number of acoustic frames and $D$ is the frame feature dimensionality, to an embedding vector $f({\bf X}) \in \mathbb{R}^d$. The goal is to learn $f$ such that segments corresponding to the same word are embedded close together, while segments of differing words are embedded farther apart.
\kledit{We use supervised training and leverage the 
labeled data available} in multiple languages besides the target language, in order to improve the performance on 
low-resource or zero-resource target languages.

\begin{figure}
    \centering
    \includegraphics[width=\linewidth]{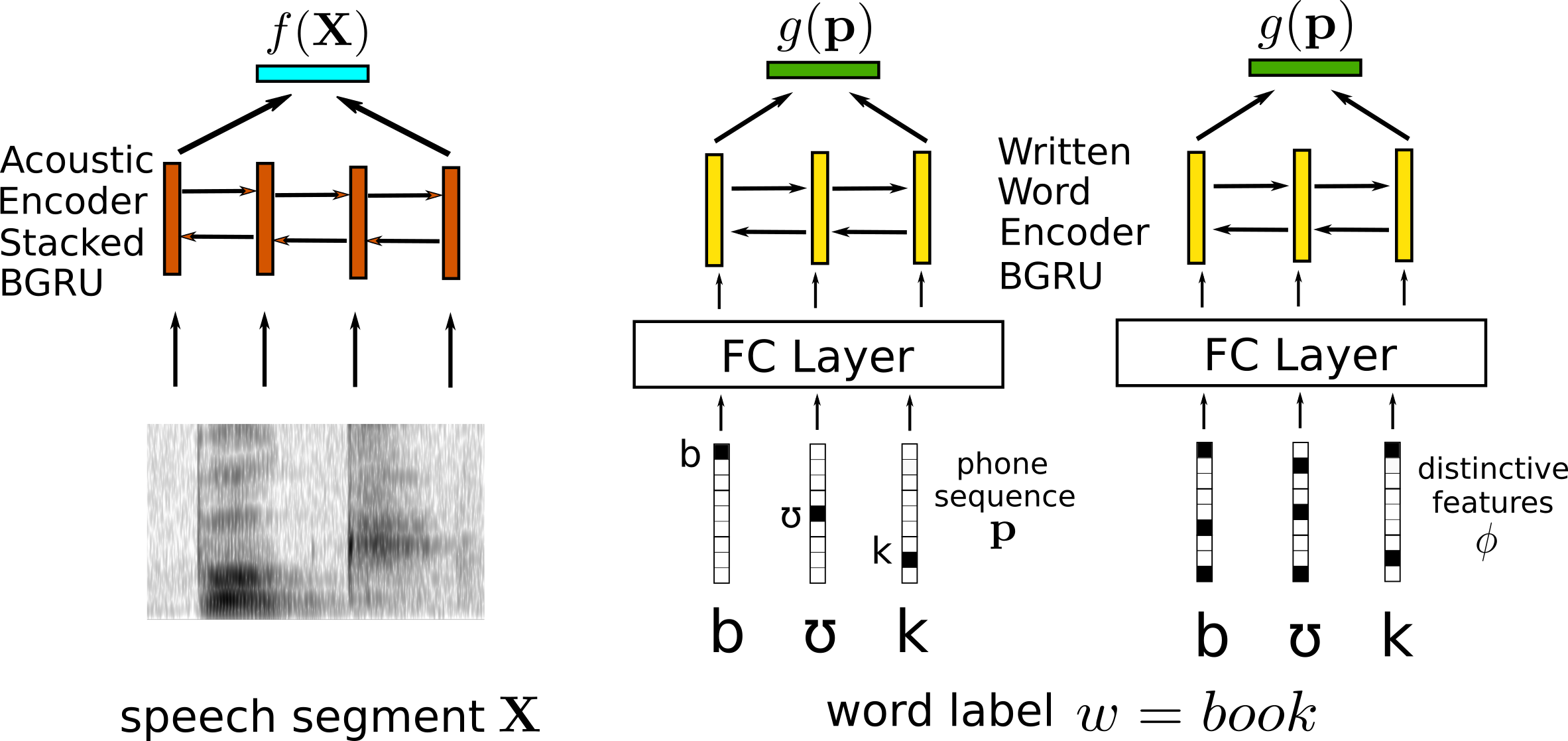}
    \caption{Acoustic word embedding (AWE) model $f$ and two acoustically grounded word embedding (AGWE) models $g$, corresponding to either phone or distinctive feature sequence input.}
    \label{fig:model}
    \vspace{-0.5cm}
\end{figure}

\vspace{-.05in}
\subsection{Embedding learning with phone-based supervision}
\vspace{-.05in}
Our approach is based on prior work on AWE/AGWE learning using a multi-view contrastive loss~\cite{he+etal_iclr2017,settle2019acoustically}, which jointly learns models for an acoustic view ($f$) and a written view ($g$) (Figure~\ref{fig:model}). 
The input to the acoustic embedding model $f$ \klcomment{we should decide if we want to use $f(), g()$ or $f,g$ and be consistent}\yhcomment{I prefer to use f and g since it is a common function notation. I changed the caption of figure 2.}\klcomment{OK -- I changed the reference to $d$ at the end of this section to match} is a variable-length spoken segment corresponding to a word. In prior work~\cite{he+etal_iclr2017,settle2019acoustically}, the input to $g$ is a character sequence, but for extension 
to multiple languages with widely differing writing systems, we instead use a phone sequence as input to $g$;
specifically, we use the Extended Speech Assessment Methods Phonetic Alphabet (X-SAMPA)~\cite{wells1995xsampa}.
We obtain the phonetic transcription from the written one using a pronunciation dictionary.  Therefore, we do not require manual phonetic transcriptions, but do require a pronunciation dictionary for each language.
Once a multilingual model is trained, however, the acoustic model $f$ can be used to embed spoken word segments in an unseen language, regardless of whether or not we have a pronunciation dictionary for it.  To further improve cross-lingual transfer, we also explore using sequences of distinctive features in place of phones (Section~\ref{sec:feats}).

\klcomment{I think we should add more indexing by language as currently the notation is not quite right.  I suggest ${\bf X}_{li}, w_{li}, {\bf p}_{li}, L_{li}$. This will make the objective expression slightly longer, but we can remove the $\min_{f,g}$ and just state that we minimize the loss. Also the second term in the loss should have a $\min$ over $X'$, not $w'$} \yhcomment{Yes agree. Notation changed}A training set for a particular language $l$ consists of pairs $\{({\bf X}_i^l, w_i^l)\}_{i=1}^{N^l}$,
where ${\bf X}_i^l \in \mathbb{R}^{T_i^l \times D}$ is a spoken word segment and $w_i^l \in \mathcal{V}^l$ is its word label. Word labels are used to determine whether two acoustic segments correspond to the same word, and to retrieve the 
phone sequence ${\bf p}_i^l \in \mathcal{P}^{L_i^l}$ for each word $w_i^l$ from a lexicon where $L_i^l$ is the phone sequence length.
The acoustic embedding model $f$ is a stacked bidirectional gated recurrent unit (BGRU) \klcomment{I think "BiGRU" may be a more common way to abbreviate this, but I may be wrong}\sscomment{I'm biased towards leaving it as is, but I'm happy to change it} network~\cite{cho2014learning}.\footnote{While previous related work has used LSTMs~\cite{he+etal_iclr2017,settle2019acoustically}, in initial experiments on English we found better performance with GRUs.} 
The embedding model $g$ of the written view consists of a learned \kledit{phonetic} embedding layer followed by a single-layer BGRU network. 
For each view, we concatenate the last time step outputs of the top layer in both directions of the BGRU as the fixed dimensional embedding.
We train $f$ and $g$ jointly to minimize the following objective: 
\vspace{-0.2cm}
\begin{flalign}
\hspace*{-0.25cm} \displaystyle \sum_{l \in \mathcal{L}} & \sum_{i=1}^{N^l}
    \left[
        m + d(f({\bf X}_{i}^{l}), g({\bf p}_{i}^{l})) - \min_{w^{\star} \in \mathcal{V}^{l} / w_{i}^{l}} d(f({\bf X}_{i}^{l}), g({\bf p}^{\star}))
    \right]_{+}&\notag\\
	+&  \left[
	    m + d(g({\bf p}_{i}^{l}), f({\bf X}_{i}^{l})) - \min_{\substack{{\bf X}^{\star} \in \\ \{{\bf X} \vert w \in \mathcal{V}^l / w_{i}^{l}\}}}
	    d(g({\bf p}_{i}^{l}), f({\bf X}^{\star}))
    \right]_{+}
\label{eq:mv}
\vspace{-0.2cm}
\end{flalign}
where $m$ is a margin hyperparameter, $d$ denotes cosine distance 
$d(u,v) = 1-\frac{u\cdot v}{\Vert u \Vert\Vert v \Vert}$, $\mathcal{L}$ is the set of training languages, $N^l$ is the number of training pairs for language $l \in \mathcal{L}$, \kledit{and ${\bf p}^{\star}$ is the phonetic pronunciation of word $w^{\star}$}. Intuitively, Equation~\ref{eq:mv} encourages embedding acoustic segments close to embeddings of their corresponding word labels and far from embeddings of other word labels, while implicitly separating acoustic segment embeddings of dissimilar words.
For efficiency, rather than finding the most offensive example across the dataset (i.e. $w^{\star} \in \mathcal{V}^l / w_i$
), we use 
\kledit{the root mean squared distance over 
the $K$ most offending} examples within the mini-batch (i.e., $\sqrt{\frac{1}{K}\sum_{k=1}^{K}{d(u, v)}^2}$ where $u,v$ represent the arguments of $d$ above). \klcomment{changed $d()$ to $d$ to match our references to $f$ and $g$}
\kledit{Each} mini-batch contains data from only one language.

\vspace{-.05in}
\subsection{Distinctive feature-based supervision}
\label{sec:feats}
\vspace{-.05in}

To make the best use of multilingual data, we would like to share as much information as possible across languages.  The X-SAMPA phone set improves over written characters, but is far from perfect, as approximately $60\%$
of phones in our 255-phone set appear in only one of the twelve languages used in our experiments. As a result, input embeddings for unseen phones are poorly (or not) learned.
To address this issue, we investigate using distinctive features (DFs), such as manner and place features, rather than phones.
\footnote{Our code, phone set, and feature set can be found at \texttt{https://github.com/Yushi-Hu/Multilingual-AWE\klcomment{fill in. I suggest you designate a page under either Yushi's or Shane's home pages and give the link here, and it's fine to take a few days to actually populate it (while Interspeech is doing reviewer assignment :)}}} 
While many phones are language-specific, all feature values we consider are used in multiple languages with the exception of click features in Zulu and tone features in Cantonese and Lithuanian. 

In this approach, we map each phone $p_j$ to a vector of its distinctive features ${\bm \phi}_j$ containing $1$ in dimensions corresponding to features that are ``on" and $0$ for features that are ``off" (see Figure~\ref{fig:model}).
We then pass this sequence of binary vectors through a linear layer which outputs vectors with the same dimensionality as the phone embeddings in the phone-based model.
This step can be viewed as computing phone embeddings as a sum of distinctive feature embeddings. After this embedding step, the rest of the model is identical to the phone-based model.

%% file: setup.tex
\vspace{-.05in}
\section{Experimental Setup}
\vspace{-.05in}

We use conversational data from 12 languages: English data from a subset of Switchboard~\cite{godfrey1992switchboard}, and 11 languages from the IARPA Babel project~\cite{babel_data}: Cantonese (31 hrs), Assamese (19 hrs), Bengali (21 hrs), Pashto (28 hrs), Turkish (32 hrs), Tagalog (30 hrs), Tamil (23 hrs), Zulu (23 hrs), Lithuanian (15 hrs), Guarani (15 hrs), and Igbo (12 hrs). The development and test sets are about 1-3 hours per language. \klcomment{do we say anywhere how many hrs in dev and test?}\yhcomment{No}\klcomment{we should -- at least a vague statement like "1-3 hours per language"}
We use  Kaldi~\cite{povey2011kaldi} \kledit{(Babel recipe \texttt{s5d})} to compute input acoustic features, train HMM/GMM triphone models, and extract word alignments for the Babel languages.
The acoustic features are $117$-dimensional
, consisting of
$36$-dimensional
log-Mel spectra + $3$-dimensional
\kledit{(Kaldi default)} pitch features. \klcomment{removed footnote and added small edits here instead.  converted "dim" to "dimensional" but we can revert if we're very low on space}
Training, development, and test sets are constructed from \kledit{non-overlapping}
conversation sides. Babel 
training sets include segments with duration $25$--$500$ frames corresponding to words occurring $\geq3$ times in training; development and test sets include segments of $50$--$500$ frames with no word frequency restrictions. English development and test sets are the same as in prior related work~\cite{carlin+etal_icassp11,levin+etal_asru13,kamper+etal_icassp15,Kamper_16a,settle2016discriminative,he+etal_iclr2017}. The English training set contains the same conversation sides as in prior work; however, for consistency within multilingual experiments we use word segments satisfying the same duration and frequency restrictions as for the Babel language training sets, which does not exactly match the prior AWE work.
For comparison with prior work on English (Table~\ref{tab:baselines}), we also train a separate set of models on the same training set as in prior work.
For distinctive feature-based experiments, the features for a given phone are retrieved from the PHOIBLE database~\cite{phoible}.

We consider the following experimental settings: {\bf single} (train and test on the target language), {\bf unseen} (train on the 11 non-target languages, and then test on the unseen target language), and {\bf fine-tune} (train on the 11 non-target languages, then fine-tune and test on the target language).
We vary the amount of training data for {\bf single} and {\bf fine-tune} experiments among 10min, 60min, and ``all" (the entire training set for the target language).
The {\bf unseen} setting is a zero-resource setting.
To tune hyperparameters in \textbf{unseen} experiments, an average evaluation score from the development sets of the 11 training languages is used such that the target language is not seen until test evaluation.

\vspace{-.05in}
\subsection{Evaluation}
\vspace{-.05in}
AWEs and AGWEs can be used for a variety of downstream tasks. Here we use a task-agnostic evaluation approach, similarly to prior work~\cite{jansen2013jhu,carlin+etal_icassp11,levin+etal_asru13,Kamper_16a,settle2016discriminative,he+etal_iclr2017}, including two ``proxy" tasks: acoustic word discrimination and cross-view word discrimination. Acoustic word discrimination is \kledit{the task of determining whether a pair of acoustic segments $({\bf X}_i, {\bf X}_j)$ correspond to the same word}, while cross-view word discrimination is \kledit{the task of determining whether an acoustic segment and word label $({\bf X}_i, w_j)$ correspond to the same word}. \klcomment{edited last 2 sentences} In both cases, we compute the cosine distance for each pair of embeddings, and consider a pair a match if its distance falls below a threshold. Results are reported as average precision (AP), i.e.~area under the precision-recall curve generated by varying the threshold. We refer to the performance measure as \kledit{``acoustic AP" for acoustic word discrimination,
and ``cross-view AP" for the cross-view task}. The acoustic word discrimination task was first introduced by~\cite{carlin+etal_icassp11} as a proxy task for query-by-example search,
and has been successfully used (along with \kledit{the cross-view task}) as a tuning criterion when applying AWEs/AGWEs to downstream \kledit{tasks}
~\cite{settle2017query,settle2019acoustically,yuan2018learning}.\yhcomment{Keith's?} \sscomment{somewhat addressed}\klcomment{somewhere in this par it should be stated that the word disc task has been found in the past to correlate well with downstream performance (I think this is in the original JHU paper), maybe also that it's been used as a tuning criterion in other work successfully applying AWEs/AGWEs to tasks}\sscomment{added Yuan's paper to list}

\vspace{-.05in}
\subsection{\kledit{Hyperparameters}}
\label{ssec:hparams}
\vspace{-.05in}
Hyperparameters are tuned on the small Switchboard subset from prior work~\cite{levin+etal_asru13,kamper+etal_icassp15,settle2016discriminative,he+etal_iclr2017,Kamper_16a}. The same model architecture is used for all languages. The acoustic view model is a 4-layer BGRU (with $0.4$ dropout rate between layers), while the written view model consists of an input embedding layer and a 1-layer BGRU. Both recurrent models use 512 hidden units per direction per layer and output 1024-dimensional embeddings. When encoding sequences of either phones or phonetic features in the written view model, the embedding layer maps input representations to 64-dimensional vectors as depicted in Figure~\ref{fig:model}. There are 255 X-SAMPA phones and 38 distinctive features. Some distinctive features can take on more than 2 values, so we represent each value of each distinctive feature separately, giving 101 learned feature \kledit{embeddings.}

During training, the margin $m$ in Equation~\ref{eq:mv} is set to $0.4$, and negative sampling uses $K=20$.
We perform mini-batch optimization with Adam~\cite{kingma2015adam}, \kledit{with batch size $256$ and initial learning rate $0.0005$}. The learning rate is decayed by a factor of $10$ if the cross-view AP on the development set(s) fails to improve over 5 epochs. Training stops when the learning rate drops below $10^{-8}$. All experiments use the PyTorch toolkit~\cite{NEURIPS2019_9015}.

%% file: results.tex
\begin{table}[t]
  \centering
  \small
  \caption{Test set performance of several embedding approaches on the English acoustic and cross-view word discrimination tasks. The numbers reported are average precision (AP).
   }
  \begin{tabular}{lcc}
    \toprule
    \textbf{Method}      & \textbf{Acoustic}  &\textbf{Cross-view}               \\
    \midrule
    \textbf{100-minute training set} \\
    $\,\,$ MFCCs + DTW \cite{Kamper_16a}                  & 0.21    \\
    $\,\,$ CAE + DTW \cite{kamper+etal_icassp15}    &  0.47 \\
    $\,\,$ Phone posteriors + DTW \cite{carlin+etal_icassp11}      &  0.50 \\
    $\,\,$ Siamese CNN \cite{Kamper_16a}&   0.55 \\
    $\,\,$ Supervised CAE-RNN~\cite{kamper2020multilingual}& 0.58\\
    $\,\,$ Siamese LSTM  \cite{settle2016discriminative}  &   0.67   \\
    $\,\,$ Multi-view LSTM~\cite{he+etal_iclr2017}~\tablefootnote{\cite{he+etal_iclr2017} calculates cross-view AP differently and is not comparable.}  &   0.81   \\
    $\,\,$ Our multi-view GRU (chars)        &    0.81 & 0.71     \\
    $\,\,$ Our multi-view GRU (phones)        & \textbf{0.84}   & \textbf{0.77}\\
    $\,\,$ Our multi-view GRU (features)  &  \textbf{0.83} &  \textbf{0.76}\\
    \midrule
    \textbf{10-hour training set} \\
    $\,\,$ Our multi-view GRU (phones)   &  \textbf{0.88}  & \textbf{0.81}\\
   $\,\,$  Our multi-view GRU (features)  & \textbf{0.87} & \textbf{0.81}\\
    \midrule
    \textbf{135-hour training set} \\
    $\,\,$ Our multi-view GRU (phones)  & \textbf{0.89} & \textbf{0.86} \\
    $\,\,$ Our multi-view GRU (features) & \textbf{0.89} & \textbf{0.86} \\
    \bottomrule
  \end{tabular}
  \vspace{-.2in}
\label{tab:baselines}
\end{table}

\vspace{-.1in}
\section{Results}
\label{sec:results}
\subsection{Comparison with prior work on English}
\vspace{-.05in}
A number of previously reported results on acoustic word embeddings have used a particular subset of Switchboard for training~\cite{levin+etal_asru13,kamper+etal_icassp15,settle2016discriminative,he+etal_iclr2017,Kamper_16a}. To compare with this work
, we also train using this same subset.
\kledit{We find (Table~\ref{tab:baselines}) that} our models 
outperform all previous methods, 
including (our implementation of) CAE-RNN from recent work on multilingual AWEs~\cite{kamper2020multilingual}.

We also find that representing the written word as a phone sequence improves over the character-based input representation used in prior work.  Based on these results, for the remaining experiments we use our multi-view GRU-based models with phonetic representations of written words.  Between the two phonetic representations (phone-based and feature-based), results are almost identical, with a slight edge for the phone-based representation.  However, for multilingual experiments, we will largely use the feature-based representation as it allows us to embed previously unseen phones, and improves \kledit{multilingual performance} (see Section~\ref{sec:df}).

Finally, we compare results of our models trained on different sizes of training set.  We find that acoustic AP plateaus by around a 10-hour training set, with the results of a model trained on only 100 minutes being not far behind.

\vspace{-.05in}
\subsection{Evaluation of multilingual acoustic word embeddings}
\vspace{-.05in}

\begin{figure}[t]
\includegraphics[width=0.99\linewidth]{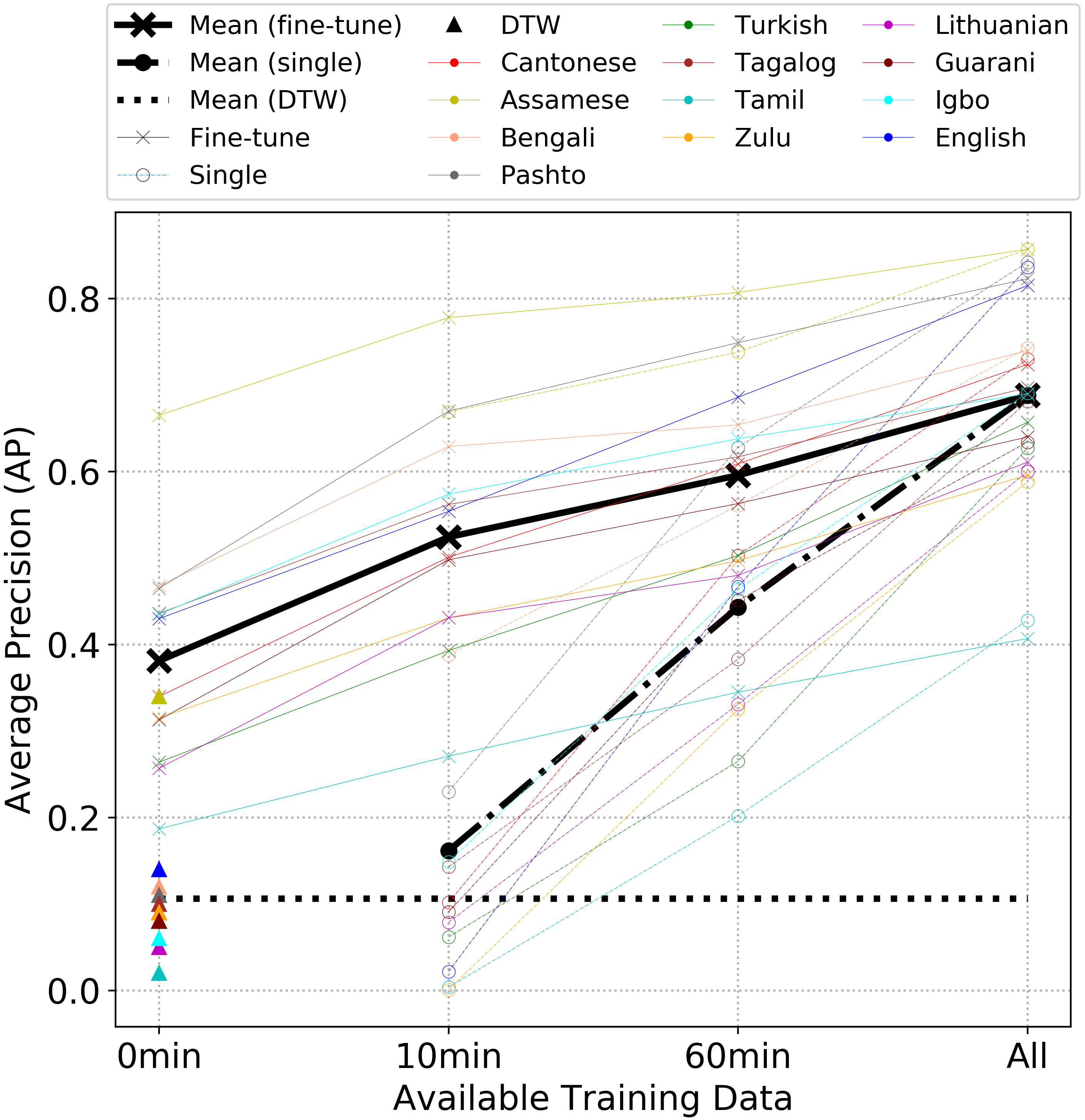}
\vspace{-0.25cm}
\caption{\kledit{Test set acoustic AP for models trained with varying amounts of target language data, supervised with distinctive features.} Note that the x-axis is not linear. ``0 min" refers to the {\bf unseen} training setting, i.e.~the {\bf fine-tune} setting with no target language training data. ``All" refers to all training data available for the target language. \kledit{Average acoustic AP across languages is given by the thick black lines; language-specific results are given by the thin colored lines.}
}
\label{fig:acoustic}
\vspace{-0.1in}
\end{figure}

 Figure~\ref{fig:acoustic} gives our main acoustic AP results for distinctive feature-based models across the 12 languages in the three training settings. (In terms of acoustic AP, there is little difference between the phone-based and distinctive feature-based models. 
These results indicate that, when resources are limited in the target language, multilingual pre-training offers clear benefits. Fine-tuning a multilingual model on 10 minutes of target language data can outperform training on 60 minutes from the target language alone.
Furthermore, if 60 minutes of data is available in the target language, multilingual pre-training cuts the performance gap between training on just that 60 minutes alone and the full $10$-$30$ hour training set by more than half.
The {\bf unseen} model (equivalent to {\bf fine-tune} with 0 minutes of target language data) is a zero-resource model with respect to the target language since it is trained on the other 11 languages and never sees examples from the target language until test evaluation. On average, our {\bf unseen} models significantly outperform the unsupervised DTW baselines---confirming results of other recent work in the zero-resource setting~\cite{kamper2020multilingual}---as well as the {\bf single}-10min models,
and perform similarly to the {\bf single}-60min models.  

\vspace{-.05in}
\subsection{Phonetic vs.~distinctive feature supervision}
\label{sec:df}
\begin{figure}[t]
  \centering
  \includegraphics[width=\linewidth]{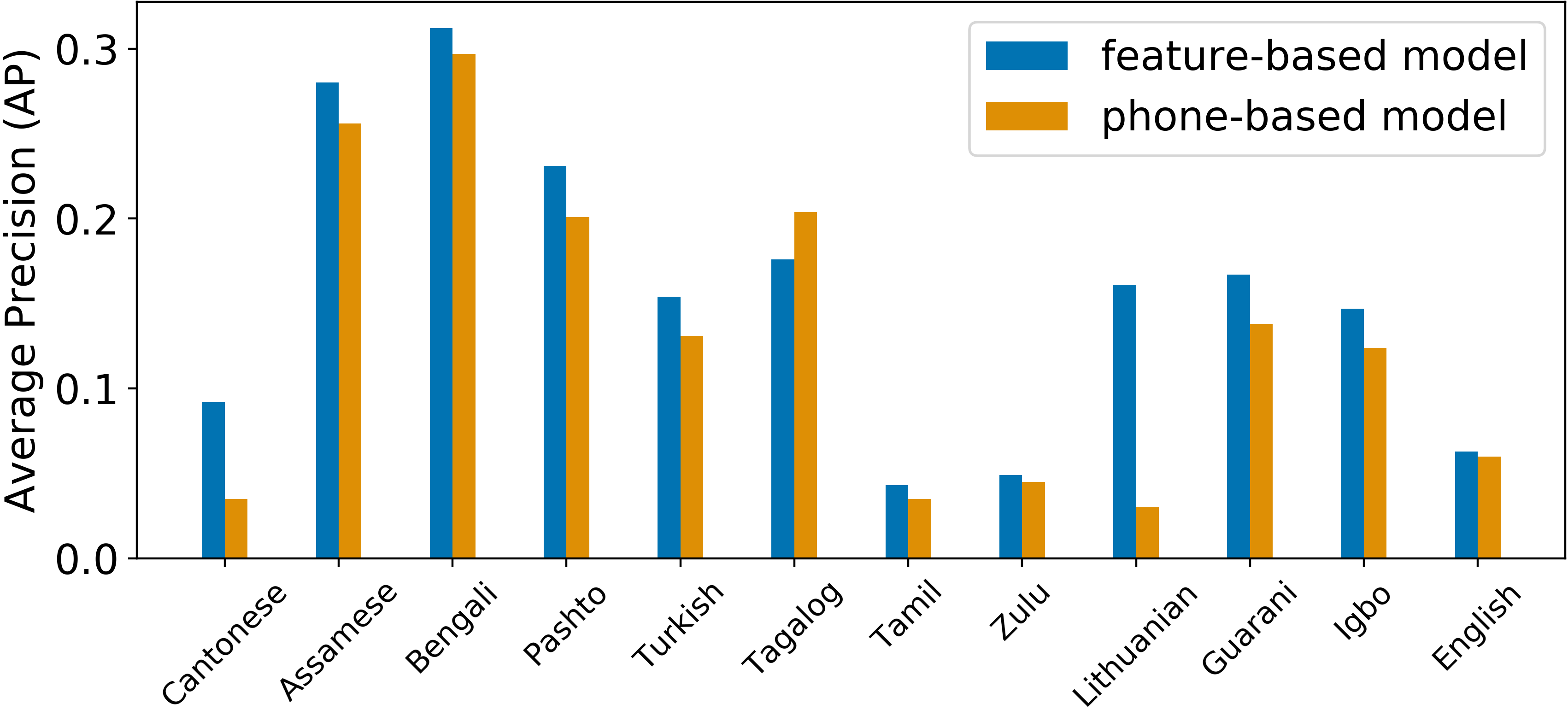}
  \caption{Test set cross-view AP in the {\bf unseen} setting.}
  \vspace{-0.35cm}
  \label{fig:cross-view}
\end{figure}

\vspace{-.05in}
\kledit{While acoustic AP is largely unaffected by the choice of phone vs.~feature supervision, in terms of cross-view AP, Figure~\ref{fig:cross-view} shows that {\bf unseen} models typically benefit from using distinctive features over phones.}  
The two languages with the largest improvement from distinctive features are Cantonese and Lithuanian. \kledit{The Cantonese data includes} a large number of diphthongs that are unseen in other languages, so their embeddings cannot be learned in the phone-based model, but the features of those diphthongs are shared with phones in other languages. 
\kledit{In the Lithuanian data, vowels are paired with their tones, making these phones unique to Lithuanian and
again making it impossible to learn the vowel embeddings from other languages using phone-based supervision}.  
All of the distinctive features, except for the Lithuanian-specific tone features themselves, are shared with other languages, making it easier for the feature-based model to learn good embeddings.

\begin{figure}[H]
  \centering
  \vspace{-0.25cm}
  \includegraphics[width=0.8\linewidth]{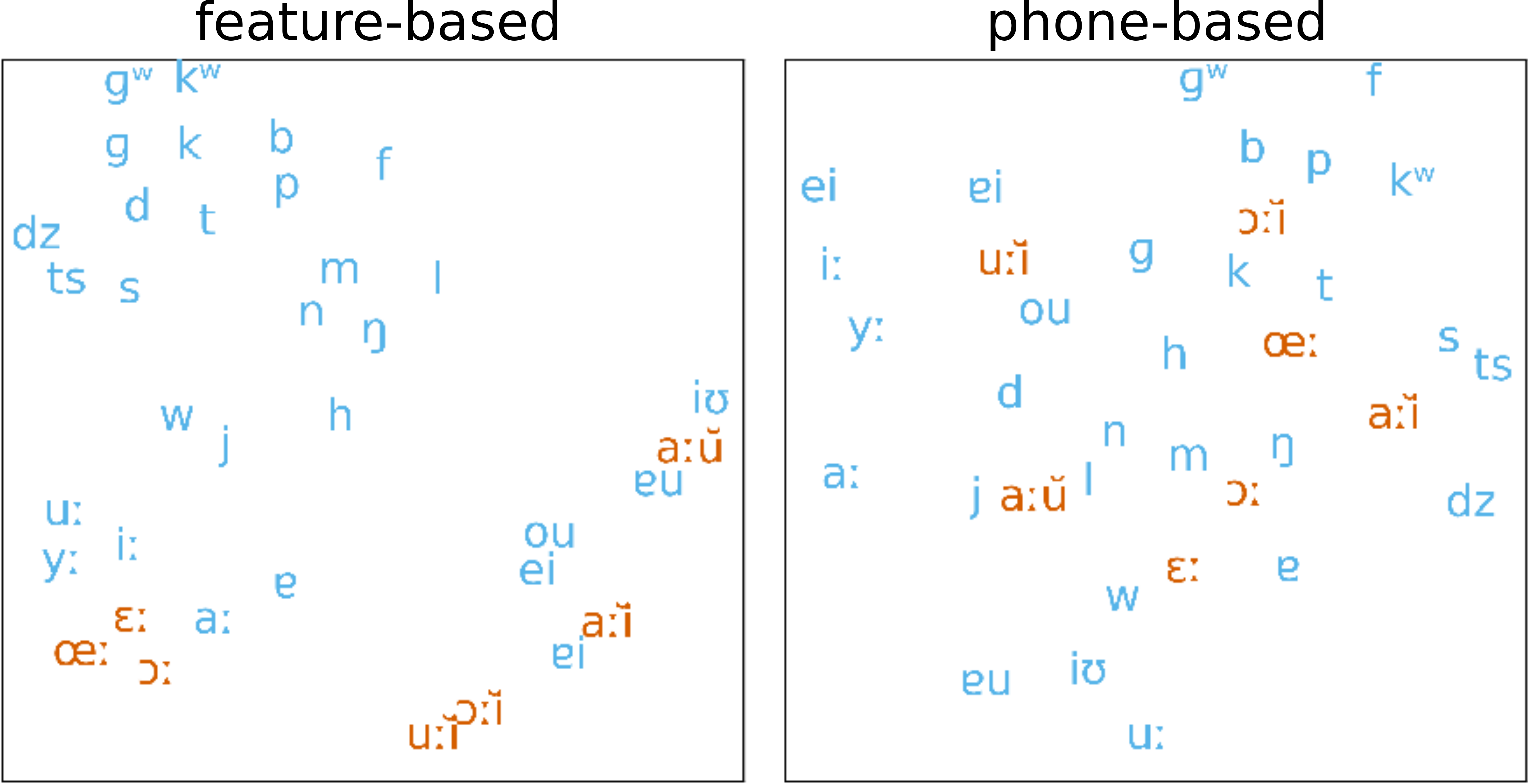}
  \caption{t-SNE~\cite{MaatenHinton08a} visualizations of Cantonese phone embeddings from \textbf{unseen} models supervised with distinctive features (left) and phones (right).
  \textcolor[rgb]{0,0.5,0.7}{Blue} phones appear in other languages; \textcolor[rgb]{0.87,0.56,0.02}{orange} phones are unique to Cantonese.}

\label{fig:cantonese-phones-tsne}
\end{figure}

In addition to generating embeddings of \kledit{spoken} and written words, our written embedding models also include a learned embedding for each phone.
Figure~\ref{fig:cantonese-phones-tsne} 
visualizes Cantonese phone embeddings taken from \kledit{a} model trained on the other 11 languages.
The model trained using distinctive features is able to infer reasonable embeddings for the phones that are unique to Cantonese and unseen in other languages, \kledit{placing them near similar phones in the embedding space,} while the phone-based model is forced to use (random) initial embeddings.

%% file: conclusions.tex
\vspace{-.1in}
\section{Conclusions}
\vspace{-.05in}
We have presented an approach for jointly learning acoustic and written word embeddings for low-resource languages, trained on data from multiple 
languages. Multilingual pre-training offers significant benefits when we have only a small amount of (or no) labeled training data for the target language.
By using distinctive features to encode the pronunciations of written words, 
we improve cross-lingual transfer by allowing phones unseen during training to share information with similar phones seen in the training set.
Future work will apply our learned embeddings to downstream tasks and expand
to a larger language set.